\documentclass[runningheads]{llncs}
\usepackage{graphicx}
\usepackage{comment}
\usepackage{color}
\usepackage[ruled,vlined,linesnumbered]{algorithm2e}
\usepackage{amsmath,amssymb} 
\usepackage{color}
\usepackage{xspace}
\usepackage[title]{appendix}
\usepackage{ctable}
\usepackage{multirow}
\usepackage{microtype}
\usepackage{amssymb}
\usepackage{pifont}
\usepackage{array, booktabs, makecell}
\usepackage[caption=false]{subfig}
\usepackage[numbers,sort&compress]{natbib}

\newcommand{\cmark}{\ding{51}}%
\newcommand{\xmark}{\ding{55}}%

\newcommand{\sceme}{\texttt{SCEME}\xspace}

\newcommand{\krish}[1]{\textcolor{purple}{Srikanth: #1}}

\newcommand{\squishlist}{
        \begin{list}{$\bullet$}
                { \setlength{\itemsep}{0pt}      \setlength{\parsep}{3pt}
                        \setlength{\topsep}{3pt}       \setlength{\partopsep}{0pt}
                        \setlength{\leftmargin}{1.0em} \setlength{\labelwidth}{1em}
                        \setlength{\labelsep}{0.5em} } }

\newcommand{\squishend}{
        \end{list}  }

\begin{document}
\pagestyle{headings}
\mainmatter
\def\ECCVSubNumber{4334}  

\title{Connecting the Dots: Detecting Adversarial Perturbations Using Context Inconsistency \thanks{This paper is accepted by ECCV 2020}}

\titlerunning{Detecting Adversarial Perturbations Using Context Inconsistency}
%
\author{Shasha Li\inst{1} \and
Shitong Zhu\inst{1} \and
Sudipta Paul\inst{1} \and \\ 
Amit Roy-Chowdhury\inst{1} \and
Chengyu Song \inst{1} \and
Srikanth Krishnamurthy\inst{1} \and \\ 
Ananthram Swami\inst{2} \and
Kevin S Chan\index{Chan, Kevin}\inst{2}}
\authorrunning{Shasha Li et al.,}
%

\institute{University of California, Riverside, USA \\ \email{\{sli057, szhu014, spaul007\}@ucr.edu, \\ \{amitrc\}@ece.ucr.edu, \{csong, krish\}@cs.ucr.edu} \and
US Army CCDC Army Research Lab \\
\email{\{ananthram.swami.civ, kevin.s.chan.civ\}@mail.mil }}
\maketitle

\frenchspacing
\begin{abstract}

There has been a recent surge in research on adversarial perturbations that
defeat Deep Neural Networks (DNNs) in machine vision; most of these perturbation-based attacks target object classifiers.
Inspired by the observation that humans are able to recognize objects that
appear out of place in a scene or along with other unlikely objects, 
we augment the DNN with a system that learns context consistency rules 
during
training and checks for the violations of the same during testing.
Our approach builds a set of auto-encoders, one for each object class, appropriately trained so as to output a discrepancy between the input and output if an added adversarial perturbation violates context consistency rules. 
Experiments on PASCAL VOC and MS COCO show that
our method effectively detects various adversarial attacks and achieves high ROC-AUC (over 0.95 in most cases);
this corresponds to over 20\% improvement over a state-of-the-art context-agnostic method.

\end{abstract}

\section{Introduction}
Recent studies have shown that
Deep Neural Networks (DNNs), which are the state-of-the-art tools for a wide range of tasks~\cite{szegedy2016rethinking,he2016deep,devlin2018bert,jin2019mmm,mccool2017mixtures},
are vulnerable to adversarial perturbation attacks~\cite{li2019stealthy,zhu2020a4}. 
In the visual domain, such adversarial perturbations can be digital or physical. 
The former refers to adding (quasi-) imperceptible digital noises to an image to cause a DNN to misclassify an object in the image; the latter refers to physically altering an object so that the captured image of that object is misclassified.
In general, adversarial perturbations are not readily noticeable by humans, but cause the machine to fail at its task. 
\begin{figure}[]
        \begin{center}
                \parbox{4.6in} {
                        \centerline{ \includegraphics[width=11cm]{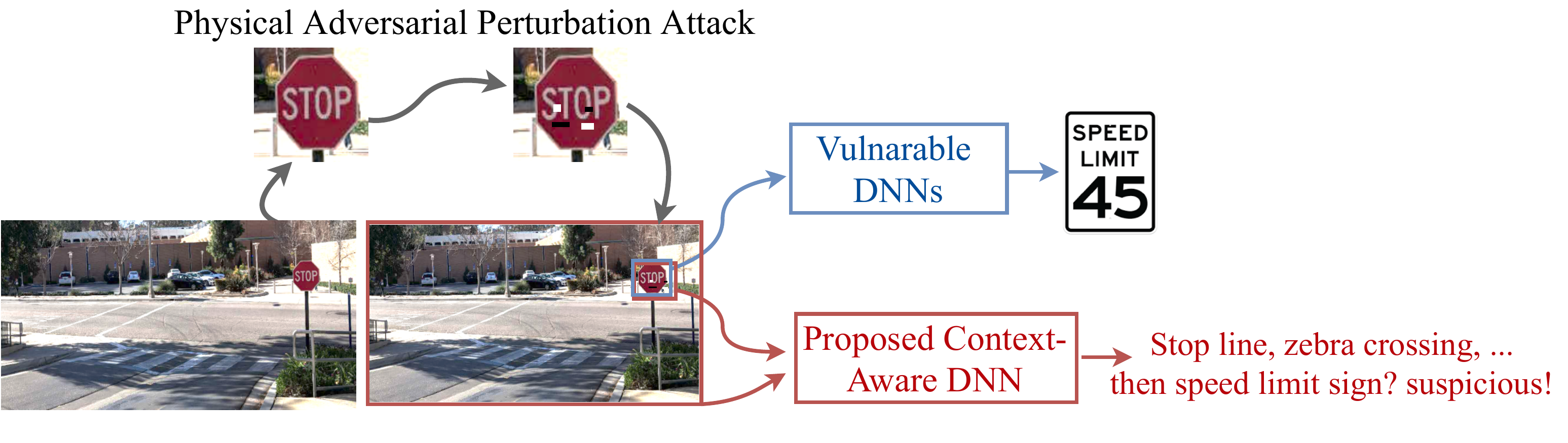}}
                        \caption{An example of how our proposed context-aware defense mechanism works. Previous studies~\cite{eykholt2018robust,song2018physical} have shown how small alterations (graffiti, patches etc.) to a stop sign make a vulnerable DNN classify it as a speed limit. We posit that a stop sign exists within the wider context of a scene (e.g., zebra crossing which is usually not seen with a speed limit sign). Thus, the scene context can be used to make the DNN more robust against such attacks.}
                        \label{fig:motivation}}
\end{center}
\end{figure} 

To defend against such attacks, our observation is that the misclassification caused by adversarial perturbations is often {\em out-of-context}.
To illustrate, consider the traffic crossing scene in Fig.~\ref{fig:motivation}; 
a stop sign often co-exists with a stop line, zebra crossing, street nameplate and other characteristics of a road intersection.
Such co-existence relationships, together with the background, create a \emph{context} that 
can be captured by human vision systems.
Specifically, if one (physically) replaces the stop sign with a speed limit sign, humans 
can recognize the anomaly that the speed limit sign does not fit in the scene.
If a DNN module can also learn such relationships (i.e., the context), it should also be able to deduce if the (mis)classification result (i.e., the speed limit sign) is out of context. 

Inspired by these observations and the fact that context has been used very successfully in recognition problems, we propose to use \emph{context inconsistency} to detect adversarial perturbation attacks.
This defense strategy complements existing defense methods~\cite{goodfellow2014explaining,jia2019comdefend,liu2019detection},
and can cope with both digital and physical perturbations.
To the best of our knowledge, it is the first strategy to defend object detection systems by considering objects ``within the context of a scene.''

\emph{We realize a system that checks for context inconsistencies caused by adversarial perturbations, and apply this approach for the defense of object detection systems};
our work is motivated by a rich literature on context-aware object recognition systems~\cite{dvornik2018modeling,barnea2019exploring,liu2018structure,hu2018relation}.
We assume a framework for object detection similar to~\cite{ren2015faster}, where the system first proposes many regions that potentially contain objects, which are then classified.
In brief, our approach accounts for four types of relationships among the regions, all of which together form the context for each proposed region:
a) regions corresponding to the same object (\textit{spatial context});
b) regions corresponding to other objects likely to co-exist within a scene (\textit{object-object context};
c) the regions likely to co-exist with the background (\textit{object-background context}); and
d) the consistency of the regions within the holistic scene (\textit{object-scene context}).
Our approach constructs a fully connected graph with the proposed regions and a super-region node which represents the scene.
In this graph, each node has, what we call an associated context profile.
The {\em context profile} is composed of node features (i.e., the original feature used for classification) and edge features (i.e., context).
Node features represent the region of interest (RoI) and edge features encode how the current region relates to other regions in its feature space representation.
Motivated by the observation that the context profile of each object category is almost always unique, we use an auto-encoder to learn the distribution of the context profile of each category.
In testing, the auto-encoder checks whether the classification result is consistent with the testing context profile.
In particular, if a proposed region (say of class $A$) contains adversarial perturbations that cause the DNN of the object detector to misclassify it as class $B$,
using the auto-encoder of class $B$ to reconstruct the testing context profile of class $A$ will result in a high reconstruction error.
Based on this, we can conclude that the classification result is suspicious.

{\color{black}The main contributions of our work are the following. \\
$\bullet$ To the best of our knowledge we are the first to propose using context inconsistency to detect adversarial perturbations in object classification tasks. \\
$\bullet$ We design and realize a DNN-based adversarial detection system that automatically extracts context for each region, and checks its consistency with a learned context distribution of the corresponding category. \\
$\bullet$ We conduct extensive experiments on both digital and physical perturbation attacks with three different adversarial targets on two large-scale datasets - PASCAL VOC~\cite{everingham2010pascal} and Microsoft COCO~\cite{lin2014microsoft}. Our method yields high detection performance in all the test cases; the ROC-AUC is over 0.95 in most cases, which is 20-35\% higher than a state-of-the-art method~\cite{xu2017feature} that does not use context in detecting adversarial perturbations.}	

\section{Related Work}
We review closely-related work and its relationship to our approach. 

\textit{Object Detection}, which seeks to locate and classify object instances in images/videos, has been extensively studied~\cite{ren2015faster,liu2016ssd,redmon2016you,lin2017focal}. Faster R-CNN~\cite{ren2015faster} is a state-of-the-art DNN-based object detector that we build upon. It initially proposes class-agnostic bounding boxes called region proposals (first stage), and then outputs the classification result for each of them in the second stage.


\textit{Adversarial Perturbations on Object Detection}, and in particular physical perturbations targeting DNN-based object detectors, have been studied recently~\cite{song2018physical,chen2018shapeshifter,zhao2019seeing} (in addition to those targeting image classifiers~\cite{kurakin2016adversarial,eykholt2018robust,athalye2017synthesizing}). 
Besides mis-categorization attacks, two new types of attacks have emerged against object detectors: the \textit{hiding attack} and the \textit{appearing attack}~\cite{chen2018shapeshifter,song2018physical}
(see Section~\ref{sec:method} for more details).
While defenses have been proposed against digital adversarial perturbations in image classification, our work focuses on both digital and physical adversarial attacks on object detection systems, which is an open and challenging problem.

\textit{Adversarial Defense} has been proposed for coping with digital perturbation attacks in the image domain. 
Detection-based defenses aim to distinguish perturbed images from normal ones. 
{\color{black}{
\textit{Statistics based detection methods} rely on extracted features that have different distributions across clean images and perturbed ones~\cite{hendrycks2016early,feinman2017detecting,liu2019detection}. 
\textit{Prediction inconsistency based detection methods} process the images and check for consistency between predictions on the original images and processed versions~\cite{xu2017feature,liang2018detecting}. 
\textit{Other methods train a second binary classifier} to distinguish perturbed inputs from clean ones~\cite{metzen2017detecting,lu2017safetynet,li2017adversarial}. 
However many of these are effective only on small and simple datasets like MNIST and 
CIFAR-10~\cite{carlini2017adversarial}. Most of them need large amounts of perturbed samples for training, and very few can be easily extended to region-level perturbation detection, which is the goal of our method. 
Table~\ref{tab:baselines} summarizes the differences between our method and 
the other defense methods; we extend
FeatureSqueeze~\cite{xu2017feature}, considered a state-of-the-art detection method, which squeezes the input features by both reducing the color bit depth of each pixel and spatially smoothening the input images, to work at the region-level and use this as a baseline {\color{black} (with this extension its performance is
directly comparable to that of our approach).}

\begin{table}[t]
\centering
\resizebox{10.0cm}{!}{
\begin{tabular}{c|c|c|c}
\specialrule{.1em}{.05em}{.05em}
Detection                       & \thead{Beyond \\ MNIST \\ CIFAR} & \thead{Do not need \\ perturbed samples \\for training} &  {\thead{Extensibility to \\object detection}} \\ \specialrule{.1em}{.05em}{.05em}
PCAWhiten~\cite{hendrycks2016early}     & \xmark                                                               & \cmark                                                                                     & \xmark, PCA is not feasible on large regions                                           \\ \hline
GaussianMix~\cite{feinman2017detecting} & \xmark                                                               & \xmark                                                                                     & \xmark, Fixed-sized inputs are required                                                 \\ \hline
Steganalysis~\cite{liu2019detection}    & \cmark                                                               & \xmark                                                                                     & \xmark, Unsatisfactory performance on small regions                              \\ \hline
ConvStat~\cite{metzen2017detecting}     & \xmark                                                               & \xmark                                                                                     & \cmark                                                                             \\ \hline
SafeNet~\cite{lu2017safetynet}          & \cmark                                                               & \xmark                                                                                     & \cmark                                                                             \\ \hline
PCAConv~\cite{li2017adversarial}        & \cmark                                                               & \xmark                                                                                     & \xmark, Fixed-sized inputs are required                                                \\ \hline
SimpleNet~\cite{gong2017adversarial}    & \xmark                                                               & \xmark                                                                                     & \cmark                                                                             \\ \hline
AdapDenoise~\cite{liang2018detecting}     & \cmark                                                               & \xmark                                                                                     & \cmark                                                                             \\ \hline
FeatureSqueeze~\cite{xu2017feature}   & \cmark                                       & \cmark                  & \cmark                                                                      \\ \specialrule{.1em}{.05em}{.05em}
\end{tabular}}
\caption{Comparison of existing detection-based defenses; since FeatureSqueeze \cite{xu2017feature} meets all the basic requirements of our approach, it is used as a baseline in the experimental analysis.}
\label{tab:baselines}
\end{table}

\textit{Context Learning for Object Detection} has been studied widely
~\cite{hollingworth1998does,oliva2003top,torralba2003contextual,dvornik2018modeling,barnea2019exploring}. 
Earlier works that incorporate context information into DNN-based object detectors~\cite{felzenszwalb2009object,choi2011tree,mottaghi2014role} use object relations in post-processing, where the detected objects are re-scored by considering object relations.
Some recent works~\cite{li2016attentive,chen2017spatial} 
perform sequential reasoning, i.e., objects detected earlier are used to help find objects later. 
The state-of-the-art approaches based on recurrent units~\cite{liu2018structure} or neural attention models~\cite{hu2018relation} process a set of objects using interactions between 
their appearance features and geometry. Our proposed context learning framework falls into this type, and among these, \cite{liu2018structure} 
is the one most related to our work.
{\color{black}{We go beyond the context learning method to define the context profile and use context inconsistency checks to detect attacks.}}

\section{Methodology}

\begin{figure}[t]
        \begin{center}
                \parbox{4.6in} {
                        \centerline{ \includegraphics[width=8cm]{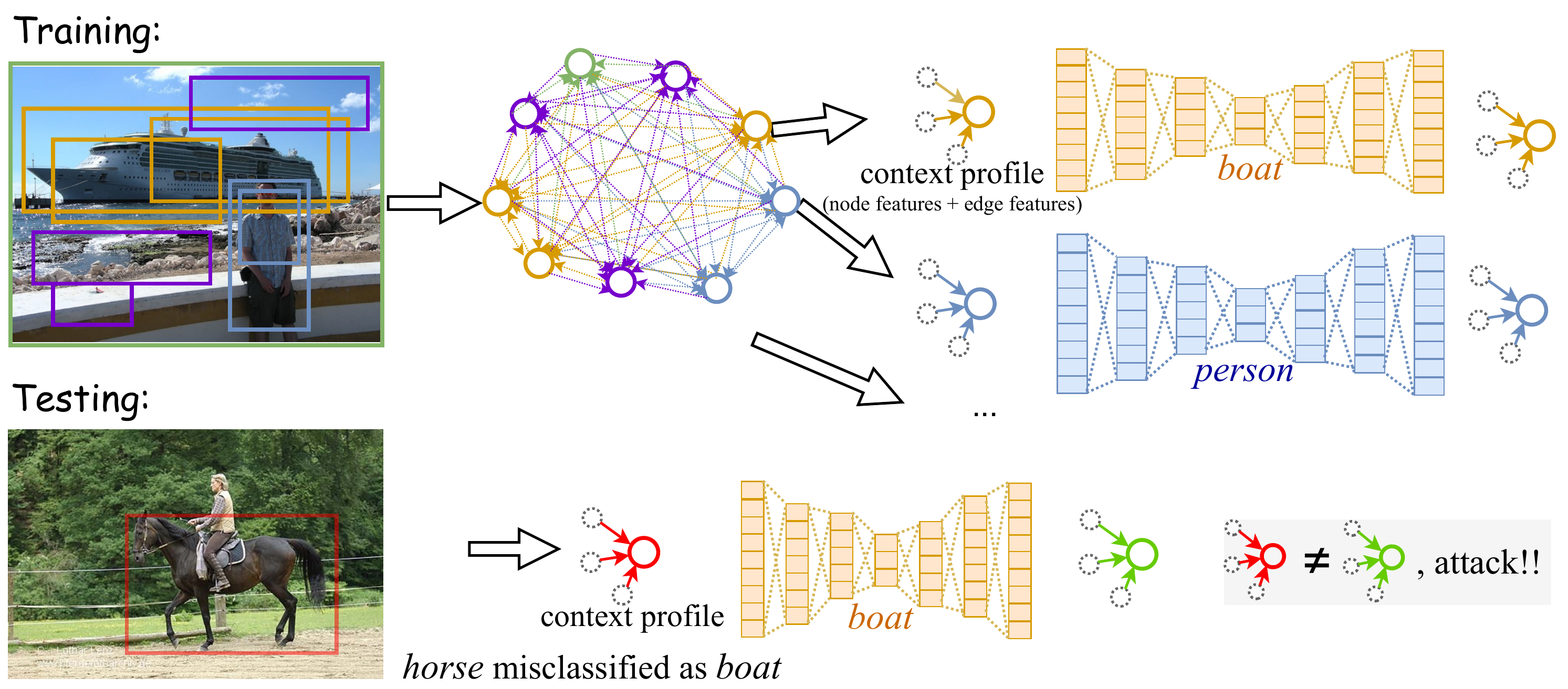}}
                    \caption{Training phase: a fully connected graph is 
built to connect the regions of the scene image -- details in Fig.~\ref{fig:SCEME}; context information relating to each object category is collected and used to train auto-encoders. Testing phase: the context profile is extracted for each region and input to the corresponding auto-encoder to check if it matches with benign distribution.}
                        \label{fig:framework}}
\end{center}
\end{figure} 
\subsection{Problem Definition and Framework Overview}
\label{sec:method}
We propose to detect adversarial perturbation attacks by recognizing the context inconsistencies they cause, i.e.,
by connecting the dots with respect to whether the object fits within the scene and in association with other entities in the scene. \\
\textbf{Threat Model.}
We assume a strong white-box attack against the two-stage Faster R-CNN model where both the training data and the parameters of the model are known to the attacker.
Since there are no existing attacks against the first stage (i.e., region proposals), we do not consider such attacks.
The attacker's goal is to cause the second stage of the object detector to malfunction by adding digital or physical perturbations to \emph{one} object instance/background region.
There are three types of attacks~\cite{song2018physical,chen2018shapeshifter,zhao2019seeing}:
\squishlist
    \item \textit{Miscategorization attacks} make the object detector miscategorize the perturbed object as belonging to a different category.
    \item \textit{Hiding attacks} make the object detector fail in recognizing the presence of the perturbed object, which happens when the confidence score is low or the object is recognized as background.
    \item \textit{Appearing attacks} make the object detector wrongly conclude that the perturbed {\color{black}background} region contains an object of a desired category.
\squishend
%
%
\textbf{Framework Overview.} We assume that we can get the region proposal results from the first stage of the Faster R-CNN model and the prediction results for each region from its second stage.
We denote the input scene image as $I$ and the region proposals as $R_I=[r_1, r_2, ..., r_N]$, where $N$ is the total number of proposals of $I$.
During the training phase, we have the ground truth category label and bounding box for each $r_i$, denoted as $S_I=[s_1, s_2, ..., s_N]$.
The Faster R-CNN's predictions on proposed regions are denoted as $\tilde{S}_I$.
Our goal as an attack detector is to identify perturbed regions from all the proposed regions.

%

Fig.~\ref{fig:framework} shows the workflow of our framework.
We use a {\color{black} structured} DNN model to build a fully connected graph on the proposed regions to model the context of a scene image.
We name this as Structure ContExt ModEl, or \sceme in short.
In \sceme, we combine the node features and edge features of each node $r_i$, to form its context profile. 
We use auto-encoders to detect context inconsistencies as outliers.
Specifically, during the training phase, for each category, we train a separate auto-encoder to capture the distribution of the benign context profile of that category.
We also have an auto-encoder for the background category to detect hiding attacks.
During testing, 
we extract the context profile for each proposed region.
We then select the corresponding auto-encoder based on the prediction result of the Faster R-CNN model and check if the testing context profile belongs to the benign distribution.
If the reconstruction error rate is higher than a threshold, we posit that the corresponding region contains adversarial perturbations.
In what follows, we describe each step of \sceme in detail.

\begin{figure}[t]
    \begin{center}
        \parbox{4.6in} {
	\centerline{ \includegraphics[width=11cm]{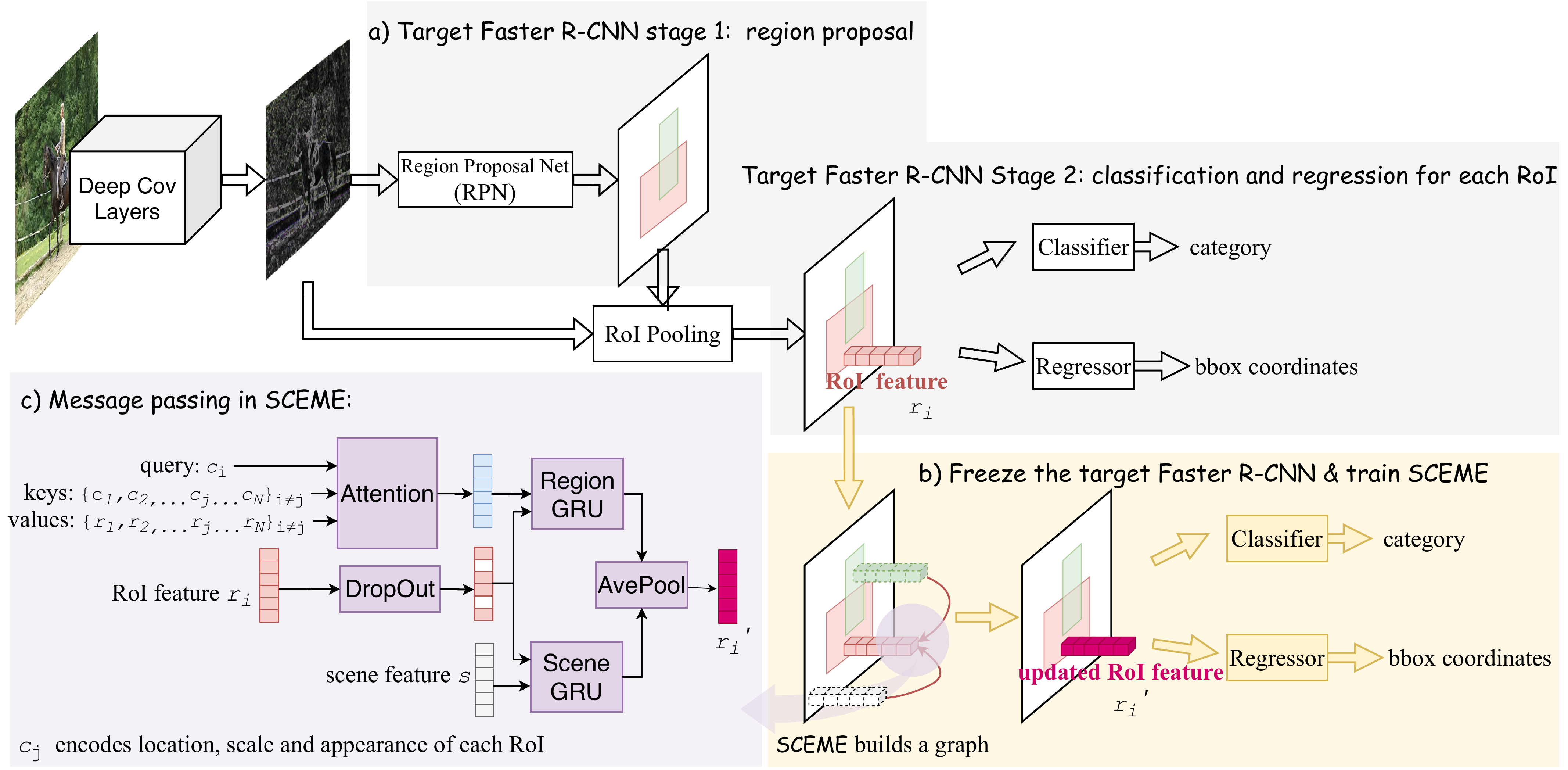}}
                \caption{\color{black}{(a) The attack target model, the Faster R-CNN, is a two-stage detector. (b) \sceme is built upon the proposed regions from the first stage of the Faster R-CNN, and updates the RoI features by message passing across regions. (c) Zooming in on SCEME shows how it fuses context information into each RoI, by updating RoI features via Region and Scene GRUs.}}
                \label{fig:SCEME}}
\end{center}
\end{figure} 

\subsection{Constructing \sceme}
In this subsection, we describe the design of the fully connected graph and the associated message passing mechanism in \sceme.
Conceptually, \sceme builds a fully connected graph on each scene image.
Each node is a region proposal generated by the first stage of the target object detector,
plus the scene node.
The initial node features, $r_i$, are the RoI pooling features of the corresponding region.
The node features are then updated ($r_i \rightarrow r_i'$) using message passing from other nodes.
After convergence, the updated node features $r_i'$ are used as inputs to a regressor towards refining the bounding box coordinates and a classifier to predict the category, as shown in Fig.~\ref{fig:SCEME}(b).
Driven by the object detection objective, we train \sceme and the following regressor and classifier together. We freeze the weights of the target Faster R-CNN during the training.
To force \sceme to rely more on context information instead of the appearance information (i.e., node features) when performing object detection, we apply a dropout function~\cite{hinton2012improving} on the node features 
before inputing into \sceme, during the training phase.
At the end of training, \sceme should be able to have better object detection performance than the target Faster R-CNN since it explicitly uses the context information from other regions to update the appearance features of each region via message passing.
This is observed in our implementation.



We use Gated Recurrent Units (GRU)~\cite{cho2014properties} with attention~\cite{bahdanau2014neural} as the message passing mechanism in \sceme. 
For each proposed region, relationships with other regions and the whole scene form four kinds of context:
\squishlist
\item {\em Same-object context}: for regions over the same object, the classification results should be consistent;
\item {\em Object-object context}: co-existence, relative location, and scale between objects are usually correlated;
\item {\em Object-background context}: the co-existence of the objects and the associated background regions are also correlated; 
\item {\em Object-scene context}: when considering the whole scene image as one super region, the co-existence of objects in the entire scene are also correlated.
\squishend
To utilize object-scene context, the scene GRU takes the scene node features $s$ as the input, and updates $r_i \rightarrow r_{scene}$.
To utilize the other kinds of context, since we have no ground truth about which object/background the regions belong to, we use attention to learn what context category to utilize from different regions.
The query and key (they encode information like location, appearance, scale, etc.) pertaining to each region are defined similar to~\cite{liu2018structure}.
Comparing the relative location, scale and co-existence between the query of the current region and the keys of all the other regions, the attention system assigns different attention scores to each region, i.e., it updates $r_i$, utilizing different amount of information from $\{ r_j\}_{j \neq i}$. Thus, $r_j$ is first weighted by the attention scores and then all $r_j$ are summed up as the input to the Region GRU to update $r_i \rightarrow r_{regions}$ as shown in Fig.~\ref{fig:SCEME}(c).
%
The corresponding output, $r_{regions}$ and $r_{scene}$, are then combined via the average pooling function to get the final updated RoI feature vector $r'$.

\subsection{Context Profile}
In this subsection, we describe how we extract a context profile in \sceme.
Recall that a context profile consists of node features $r$ and edge features, where the edge features describe how $r$ is updated.
Before introducing the edge features that we use, we describe in detail how message passing is done with GRU~\cite{cho2014properties}.

A GRU is a memory cell that can remember the initial node features $r$ and then fuse incoming messages from other nodes into a meaningful representation.
Let us consider the GRU that takes the feature vector $v$ (from other nodes) as the input, and updates the current node features $r$.
Note that $r$ and $v$ have the same dimensions since both are from RoI pooling.
GRU computes two gates given $v$ and $r$, for message fusion. 
The reset gate $\gamma_r$ 
drops or enhances information in the initial memory based on its relevance to the incoming message $v$.
The update gate $\gamma_u$ 
controls how much of the initial memory needs to be carried over to the next memory state, thus allowing a more effective representation.
%
In other words, $\gamma_r$ and $\gamma_u$ are two vectors of the same dimension as $r$ and $v$, which are learned by the model to decide what information should be passed to the next memory state given the current memory state and the incoming message.
Therefore, we use the gate vectors as the edge features in the context profile. There are, in total, four gate feature vectors from both the Scene GRU and the Region GRU.
%
Therefore, we define the context profile of a proposed region as $x=[r,\gamma_{u1},\gamma_{u2},\gamma_{r1},\gamma_{r2}]$.

\begin{algorithm}[t]
	\fontsize{8}{8}
	\selectfont
	\SetAlgoLined
	\SetKwInOut{Input}{Input}\SetKwInOut{Output}{Output}
	\SetKwFunction{CP}{ExtractContextProfiles}
	\SetKwFunction{TrainSCEME}{TrainSCEME}
	\SetKwFunction{GetPred}{GetPredictedCategory}
	\SetKwFunction{TrainAuto}{TrainAutoEncoder}
	\SetKwFunction{Thresh}{GetErrThreshold}
	
	\Input{ $\{R_I, S_I, \tilde{S}_I\}_{I \in TrainSet}$}
	\Output{\sceme, $AutoEncoder_{c}$ for each object category $c$, and $thresh_{err}$}
	
	\sceme $\gets$ \TrainSCEME{ $\{R_I, S_I\}_{I \in TrainSet}$ }\\
	$\text{ContextProfiles}[c] = []$ for each  object category c
	
	\For {each $R_I = [r_1, r_2, ...] $}{
		$X_I = [x_1, x_2,...] \gets $ \CP{\sceme, $R_I$}\\
		\For {each region, its prediction, and its context profile $ \{r_j,\tilde{s_j}, x_j\} $  }{
			$\tilde{c} \gets $ \GetPred{$\tilde{s_j}$} \\
			$\text{ContextProfiles}[\tilde{c}] \gets \text{ContextProfiles}[\tilde{c}] + x_j$
		}	
	}
	\For {each category $c$}{
		$AutoEncoder_{c} \gets $ \TrainAuto{$\text{ContextProfiles}[c]$} 
	}
	$thresh_{err} = $ \Thresh{$\{AutoEncoder_{c}\}$}\\
	\Return \sceme, $\{ AutoEncoder_{c}\}$, $thresh_{err}$
	\caption{\sceme: Training phase}
	\label{algo:training}
\end{algorithm}

\subsection{AutoEncoder for Learning Context Profile Distribution}

In benign settings, all context profiles of a given category must be similar to each other. For example, {\color{black} stop sign features exist with features of road signs} and zebra crossings.
Therefore, the context profile of a stop sign corresponds to a unique distribution that accounts for these characteristics.
When a stop sign is misclassified as a speed limit sign, its context profile should not fit with the distribution corresponding to that of the speed limit sign category.

For each category, we use a separate auto-encoder 
(architecture shown in the supplementary material)
to learn the distribution of its context profile.
The input to the auto-encoder is the context profile $x=[r,\gamma_{u1},\gamma_{u2},\gamma_{r1},\gamma_{r2}]$.
A fully connected layer is first used to compress the node features ($r$) and edge features ($[\gamma_{u1},\gamma_{u2},\gamma_{r1},\gamma_{r2}]$) separately.
This is followed by two convolution layers, wherein the node and edge features are combined to learn the joint compression.
Two fully connected layers are then used to further compress the joint features.
These layers form a bottleneck that drives the encoder to learn the true relationships between the features and get rid of redundant information.
{\textcolor{black}SmoothL1Loss, as defined in ~\cite{huber1992robust,xie2020high}, between the input and the output is used to train the auto-encoder, which is a common practice.}

Once trained, we can detect adversarial perturbation attacks by appropriately thresholding the reconstruction error.
Giving a new context profile during testing, 
if a) the node features are not aligned with the corresponding distribution of benign node features, or
b) the edge features are not aligned with the corresponding distribution of benign edge features, or 
c) the joint distribution between the node features and the edge features is violated, 
the auto-encoder will not be able to reconstruct the features using its learned distribution/relation. 
In other words,  a reconstruction error that is larger than the chosen threshold would indicate either an appearance discrepancy or a context discrepancy between the input and output of the auto-encoder.

An overview of the approach (training and testing phases) is captured in Algorithms \ref{algo:training} and \ref{algo:testing}.

\begin{algorithm}[t]
	\fontsize{8}{8}
	\selectfont
	\SetAlgoLined
	\SetKwInOut{Input}{Input}\SetKwInOut{Output}{Output}
	\SetKwFunction{CP}{ExtractContextProfiles}
	\SetKwFunction{Err}{GetAutoEncoderReconErr}
	\SetKwFunction{Region}{GetRegion}
	\Input{$R_I$, $\tilde{S}_I$, \sceme, $\{AutoEncoder_{c}\}$, $thresh_{err}$}
	\Output{perturbed regions $PerturbedSet$}
	$PerturbedSet = []$\\
	$X_I =$ \CP{\sceme, $R_I$}\\
	\For {each region, its prediction, and its context profile $ \{r_j, \tilde{s_j}, x_j\} $  }{
		$\tilde{c} \gets $ \GetPred{$\tilde{s_j}$} \\
		err = \Err{$AutoEncoder_{\tilde{c}}, x_j$}\\
		\uIf{$err > thresh_{err}$}{
			region $\gets$ \Region{$\tilde{s_j}$}\\
			$PerturbedSet \gets PerturbedSet + $ region
		}
	}
	\Return $PerturbedSet$
	\caption{\sceme: Testing phase}
	\label{algo:testing}
\end{algorithm}

\section{Experimental Analysis}
We conduct comprehensive experiments on two large-scale object detection datasets to evaluate the proposed method, \sceme, against six different adversarial attacks, viz., digital miscategorization attack, digital hiding attack, digital appearing attack, physical miscategorization attack, physical hiding attack, and physical appearing attack, on Faster R-CNN (the general idea can be applied more broadly). We analyze how different kinds of context contribute to the detection performance. We also provide a case study for detecting physical perturbations on stop signs, which has been used widely as a motivating example.

\subsection{Implementation Details}

\noindent \textbf{Datasets.}
We use both PASCAL VOC~\cite{everingham2010pascal} and MS COCO~\cite{lin2014microsoft}. 
PASCAL VOC contains 20 object categories.
Each image, on average, has 1.4 categories and 2.3 instances~\cite{lin2014microsoft}. We use \textit{voc07trainval} and \textit{voc12trainval} as training datasets and the evaluations are carried out on \textit{voc07test}. 
MS COCO contains 80 categories. Each image, on average, has 3.5 categories and 7.7 instances. \textit{coco14train} and \textit{coco14valminusminival} are used for training, and the evaluations are carried out on \textit{coco14minival}.
Note that COCO has few examples for certain categories. 
To make sure we have enough number of context profiles to learn the distribution, we train 11 auto-encoders for {\color{black} the 11 categories that have the largest numbers of 
extracted context profiles}. Details are provided in the supplementary material.

\noindent \textbf{Attack Implementations.}
For digital attacks, we use the standard iterative fast gradient sign method (IFGSM)~\cite{kurakin2016adversarial} and constrain the perturbation location within the ground truth bounding box of the object instance.
Because our defense depends on contextual information, it is not sensitive to how the perturbation is generated.
We compare the performance against perturbations generated by a different method (FGSM) in the supplementary material.  
We use the physical attacks proposed in~\cite{eykholt2018robust,song2018physical}, where 
perturbation stickers are constrained to be on the object surface; the color of the stickers should be printable, and the pattern of the stickers should be smooth. For evaluations on a large scale, we do not print or add stickers physically; we add them digitally onto the scene image. This favors attackers since they can control how their physical perturbations are captured. 

\noindent \textbf{Defense Implementation.}
Momentum optimizer with momentum 0.9 is used to train \sceme. The learning rate is 5e-4 and decays every 80k iterations at a decay rate of 0.1. The training finishes after 250k iterations.
Adam optimizer is used to train auto-encoders. The learning rate is 1e-4 and reduced by 0.1 when the training loss stops decreasing for 2 epochs. Training finishes after 10 epochs.

\begin{figure}[t]
	\centering
	\subfloat[]{
		\label{fig:recon_error}
		\includegraphics[width=0.48\textwidth]{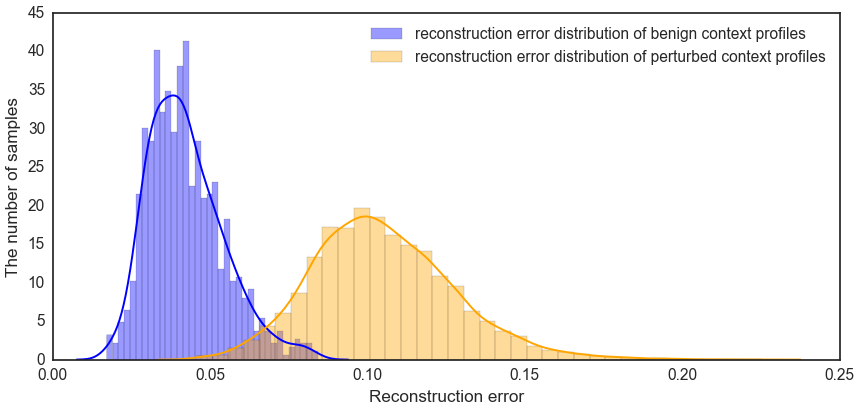} } 
	\hfill
	\subfloat[]{
		\label{images/roc_curves.pdf}
		\includegraphics[width=0.48\textwidth]{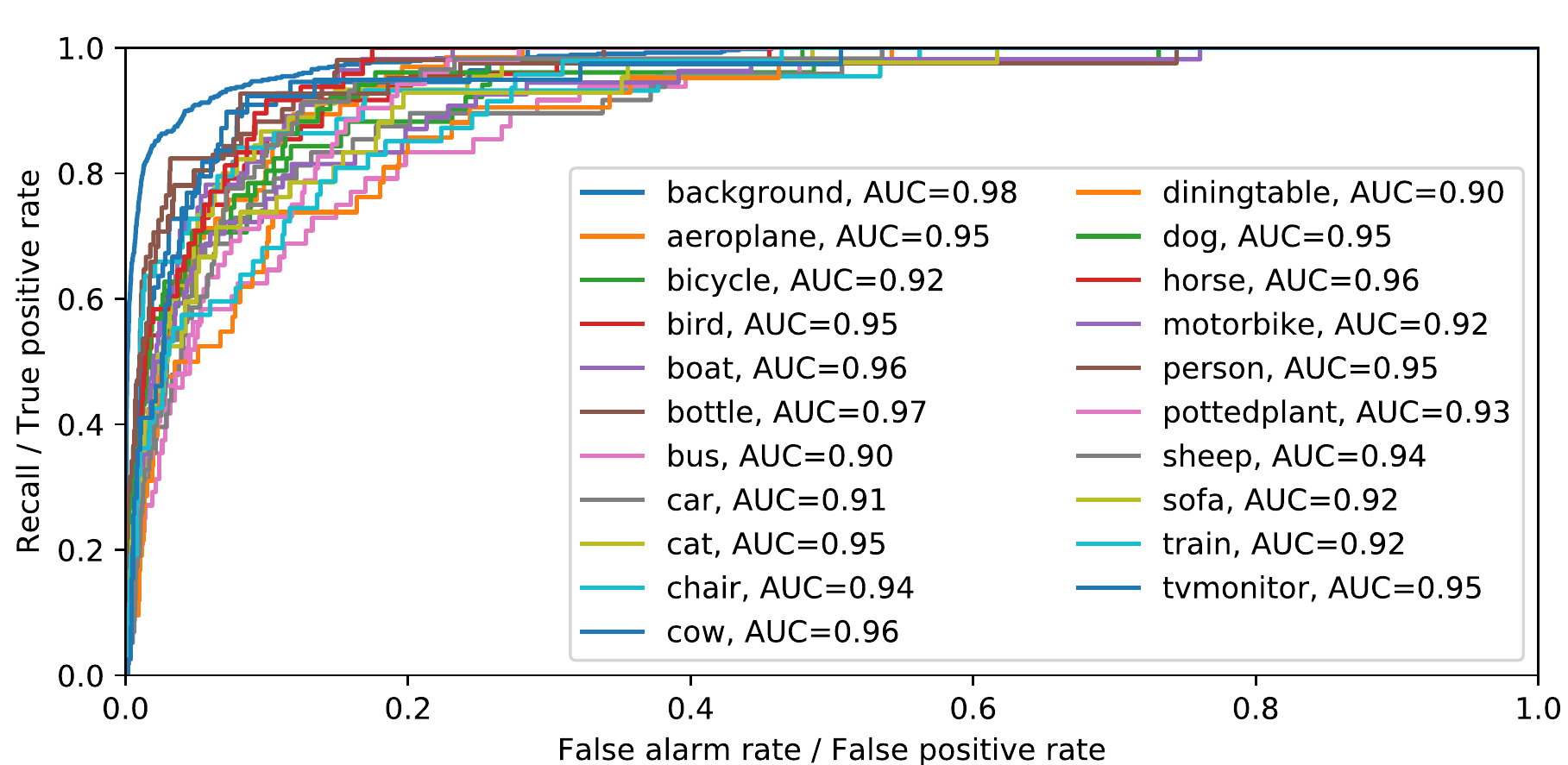} } 
	\caption{(a) Reconstruction errors of benign aeroplane context profiles are generally smaller than those of the context profiles of digitally perturbed objects that are misclassified as an aeroplane. (b) Thresholding the reconstruction error, we get the detection ROC curves for all the categories on PASCAL VOC dataset.} 
	\label{fig:recon_error} 
\end{figure}

\subsection{Evaluation of Detection Performance}

\noindent \textbf{Evaluation Metric.} We extract the context profile for each proposed region, feed it to its corresponding auto-encoder and threshold the reconstruction error to detect adversarial perturbations. Therefore, we evaluate the detection performance at the region level. Benign/negative regions are the regions proposed from clean {\color{black} objects}; perturbed/positive regions are the regions relating to perturbed objects.
We report Area Under Curve (AUC) of Receiver Operating Characteristic Curve (ROC) to evaluate the detection performance. 
Note that there can be multiple regions of a perturbed object. If any of these regions is detected, it is a successful perturbation detection. For hiding attacks, there is a possibility 
of no proposed region; 
however, it occurs rarely (less than 1\%).

\noindent \textbf{Visualizing the Reconstruction Error.} We plot the reconstruction error of benign aeroplane context profiles and that of digitally perturbed objects that are misclassified as an aeroplane. As shown in Fig.~\ref{fig:recon_error}(a), the context profiles of perturbed regions do not 
conform with the benign distribution of aeroplanes' context profiles and cause larger reconstruction errors. This test validates our hypothesis that the context profile of each category has a unique distribution. The auto-encoder that learns from the context profile of class A will not reconstruct class B well.

\begin{table}[t]
	\centering
	\resizebox{10.5cm}{!}{
		\begin{tabular}{ccccccc}
			\specialrule{.1em}{.05em}{.05em}
			\multirow{2}{*}{Method}  & \multicolumn{3}{c}{Digital Perturbation} & \multicolumn{3}{c}{Physical Perturbation} \\  
			& Miscateg  & Hiding  & Appearing & Miscateg  & Hiding  & Appearing  \\ \specialrule{.1em}{.05em}{.05em}
			\textbf{\footnotesize{Results on PASCAL VOC}:} & & & & & & \\
			\specialrule{.1em}{.05em}{.05em}
			FeatureSqueeze~\cite{xu2017feature}           & 0.724              & 0.620   & 0.597     & 0.779              & 0.661   & 0.653      \\ \hline
			Co-occurGraph~\cite{bappy2016online}  & 0.675              & -       & -         & 0.810              & -       & -          \\ \hline
			\sceme (node features only) & 0.866              & 0.976   & 0.828     & 0.947              & 0.964   & 0.927      \\ \hline
			\sceme                     & 0.938              & 0.981   & 0.869     & 0.973              & 0.976   & 0.970      \\ \specialrule{.1em}{.05em}{.05em}
			\textbf{\footnotesize{Results on MS COCO}:} & & & & & & \\
			\specialrule{.1em}{.05em}{.05em}
			FeatureSqueeze~\cite{xu2017feature}           & 0.681              & 0.682   & 0.578     & 0.699              & 0.687   & 0.540      \\ \hline
			Co-occurGraph~\cite{bappy2016online}  & 0.605              & -       & -         & 0.546             & -       & -          \\ \hline
			\sceme (node features only) & 0.901              & 0.976   & 0.810     & 0.972              & 0.954   & 0.971      \\ \hline
			\sceme                     & 0.959              & 0.984   & 0.886     & 0.989              & 0.968   & 0.989      \\ \specialrule{.1em}{.05em}{.05em}
	\end{tabular}}
	\caption{The detection performance (ROC-AUC) against six different attacks on PASCAL VOC and MS COCO dataset}
	\label{tab:detection_performance}
\end{table}

\noindent \textbf{Detection Performance.}
Thresholding the reconstruction error, we plot the ROC curve for ``aeroplane'' and other object categories tested on PASCAL VOC dataset, in Fig.~\ref{fig:recon_error}(b). 
The AUCs for all 21 categories (including background) are all over 90\%. This means 
that all the categories have their unique context profile distributions, and the reconstruction error of their auto-encoders effectively detect perturbations. 
The detection performance results, against six attacks on PASCAL VOC and MS COCO, are shown in Tab.~\ref{tab:detection_performance}. 
Three baselines are considered.
\squishlist
	\item {\em FeatureSqueeze}~\cite{xu2017feature}. As discussed in Tab.~\ref{tab:baselines}, many existing adversarial perturbation detection methods are not effective beyond simple datasets. Most require perturbed samples while training, and only few can be extended to region-level perturbation detection. We extend FeatureSqueeze, one of the state-of-the-art methods, {\color{black} that is not limited by these}, for the object detection task. Implementation details are provided in the supplementary material. 		
 	\item {\em Co-occurGraph}~\cite{bappy2016online}. We also consider a non-deep graph model where co-occurrence context is represented, as a baseline. We check the inconsistency between the relational information in the training data and testing images to detect attacks. Details are in the supplementary material. Note that the co-occurrence statistics of background class cannot be modeled, and so this approach is inapplicable for detecting hiding and appearing attacks.  	
 	\item {\em \sceme (node features only)}. Only node features are used to train the auto-encoders (instead of using context profiles with both node features for region representation and edge features for contextual relation representation). Note that the node features already implicitly contain context information since, with Faster R-CNN, the receptive field of neurons grows with depth and eventually covers the entire image. We use this baseline to quantify the improvement we achieve by explicitly modeling context information with \sceme .
\squishend

Our method \sceme, yields high AUC on both datasets and for all six attacks; many of them are over 0.95. The detection performance of \sceme is consistently better than that of FeatureSqueeze, by over 20\%. Compared to Co-occurGraph, the performance of our method in detecting miscategorization attacks, is better by over 15\%. Importantly, \sceme is able to detect hiding and appearing attacks and detect perturbations in images with one object, which is not feasible with Co-occurGraph. Using node features yields good detection performance and further using edge features, improves performance by up to 8\% for some attacks. 
\begin{figure}[t]
        \begin{center}
                \parbox{4.6in} {
                        \centerline{ \includegraphics[width=10.5cm]{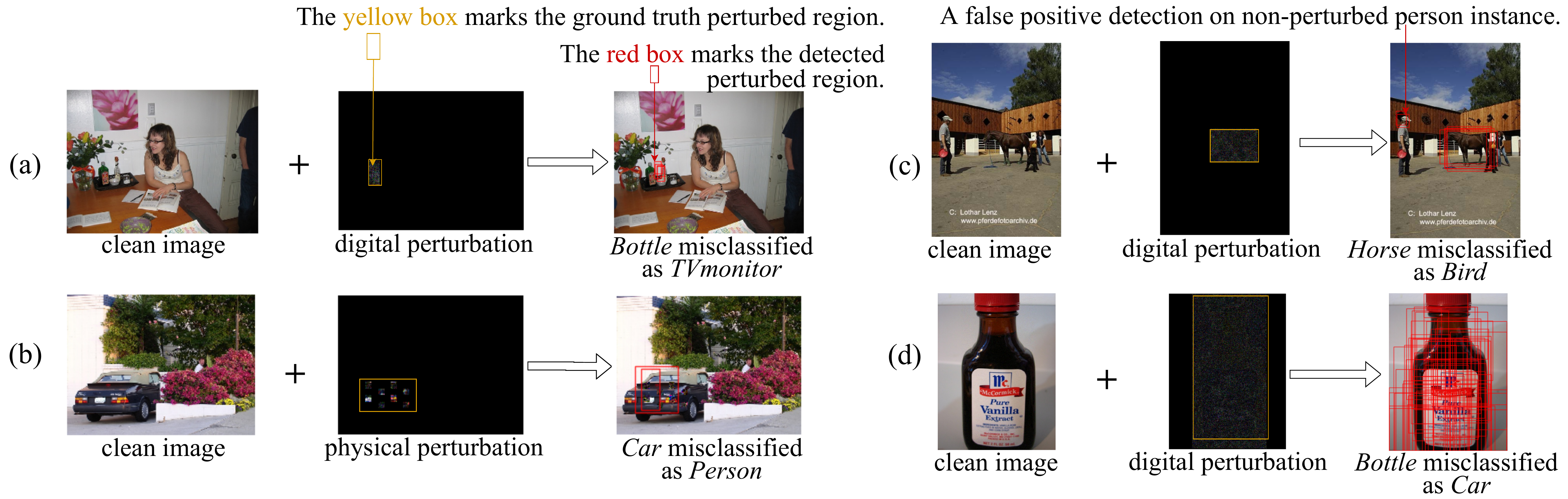}}
                        \caption{ A few interesting examples.
                        {\color{black}{\sceme successfully detects both digital and physical perturbations as shown in (a) and (b). (c) shows that the horse misclassification affects the context profile of person and leads to false positive detection on the person instance. (d) Appearance information and spatial context are used to successfully detect perturbations. }}
                        }                        \label{fig:visual}}
\end{center}
\end{figure}

\noindent \textbf{Examples of Detection Results.} We visualize the detected perturbed regions for both digital and physical miscategorization attack in Fig.~\ref{fig:visual}. 
The reconstruction error threshold is chosen to make the false positive rate 0.2\%. \sceme successfully detects both digital and physical perturbations as shown in Fig.~\ref{fig:visual}(a)and(b). {\color{black}{The misclassification of the perturbed object could affect the context information of another coexisting benign object and lead to a false perturbation detection on the benign object}} as shown in Fig.~\ref{fig:visual}(c). 
We observe that this rarely happens. In most cases, although some part of the object-object context gets violated, the appearance representation and other context would help in making the right detection. When there are not many object-object context relationships as shown in Fig.~\ref{fig:visual}(d), appearance information and spatial context are mainly used to detect a perturbation.

\begin{figure}[t]
        \begin{center}
                \parbox{4.6in} {
                        \centerline{ \includegraphics[width=11.5cm]{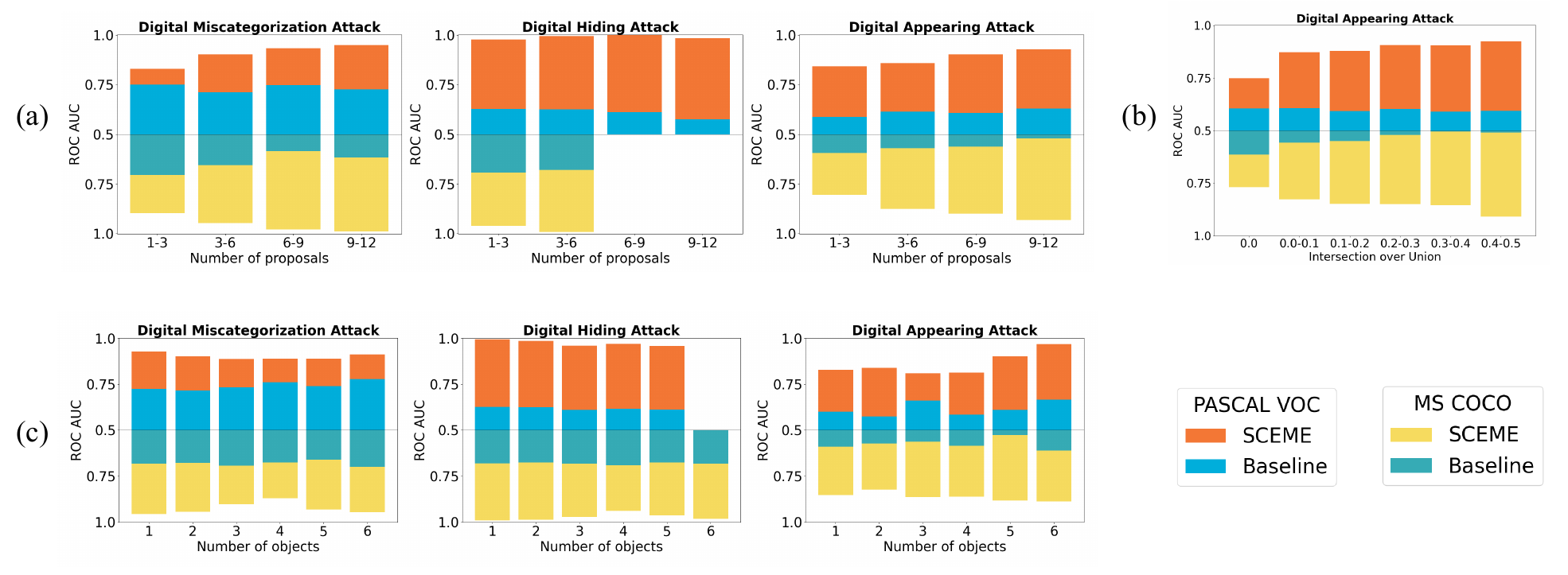}}
                        \caption{{\color{black}{Subfigures are diverging bar charts. They start with ROC-AUC = 0.5 and diverge in both upper and lower directions: upper parts are results on PASCAL VOC and lower parts are on MS COCO. For each dataset, we show both the results from the FeatureSqueeze baseline and \sceme, using overlay bars.}} 
		(a) The more the regions proposed, the better our detection performs, as there is more utilizable spatial context;
		(b) the larger the overlapped region between the ``appearing object'' and another object, the better our detection performs, as the spatial context violation becomes larger and detectable (we only analyze the appearing attack here);
		(c) the more the objects, the better our detection performs generally, as there is more utilizable object-object context (performance slightly saturates at first due to inadequate spatial context).
}
                            \label{fig:analysis}}
\vspace{-0.45in}
\end{center}
\end{figure}
\subsection{Analysis of Different Contextual Relations}
In this subsection, we analyze what roles different kinds of context features play. 

\noindent \textbf{Spatial context} consistency means that nearby regions of the same object should yield consistent prediction. 
We do two kinds of analysis. The first one is to observe the correlations between the adversarial detection performance and the number of regions proposed by the target Faster R-CNN for the perturbed object.
Fig.~\ref{fig:analysis}(a) shows that the detection performance improves when more regions are proposed for the object and this correlation is not observed for the baseline method (for both datasets). This indicates that spatial context plays a role in perturbation detection. 
Our second analysis is on appearing attacks. If the ``appearing object'' has a large overlap with one ground truth object, the spatial context of that region will be violated. We plot in Fig.~\ref{fig:analysis}(b) 
the detection performance with respect to the overlap between the appearing object and the ground truth object, measured by Intersection over Union (IoU). We observe that the more these two objects overlap, the more likely the region is detected as perturbed, consistent with our hypothesis.

\noindent \textbf{Object-object context} captures the co-existence of objects and their relative position and scale relations. We test the detection performance with respect to the number of objects in the scene images. As shown in Fig.~\ref{fig:analysis}(c), in most cases, the detection performance of \sceme first drops or stays stable, and then improves. We believe that the reason is as follows: initially, as the number of objects increases, the object-object context is weak and so is the spatial context as the size of the objects gets smaller with more of them; however, as the number of objects increases, the object-object context dominates and performance improves.

\subsection{Case Study on Stop Sign}
\label{sec:stop_sign}
We revisit the stop sign example and provide quantitative results to validate that 
context information helps defend against perturbations. We get 1000 perturbed stop sign examples, all of which are misclassified by the Faster RCNN, from the COCO dataset. 
The baselines and \sceme, are tested for detecting the perturbations. If we set a lower reconstruction error threshold, we will have a better chance of detecting the perturbed stop signs. However, there will be higher false positives, which means wrong categorization of clean regions as perturbed. Thus, to compare the methods, we constrain the threshold of each method 
so as to meet a certain \textit{False Positive Rate} (FPR), and compute the \textit{recall} 
achieved, i.e., out of the 1000 samples, how many are detected as perturbed?
The results are shown in Tab.~\ref{tab:stop_sign}. FeatureSqueeze~\cite{xu2017feature} cannot detect any perturbation until a FPR 5\% is chosen. \sceme detects 54\% of the perturbed stop signs with a FPR of 0.1\%. 
{\color{black}
Further, compared to its ablated version (that only uses node features), our method detects almost twice as many perturbed samples when the FPR required is very low (which is the case in many real-world applications).
}

\begin{table}[t]
\centering
\setlength{\tabcolsep}{3mm}{
\resizebox{10.5cm}{!}{
\begin{tabular}{c|c|c|c|c|c}
\specialrule{.1em}{.05em}{.05em}
False Positive Rate         & 0.1\% & 0.5\% & 1\%  & 5\%  & 10\% \\ \specialrule{.1em}{.05em}{.05em}
Recall of FeatureSqueeze~\cite{xu2017feature} & 0     & 0     & 0    & 3\%  & 8\%  \\ \hline
Recall of \sceme (node features only)  & 33\%     & 52\%     & 64\%    & 83\%  & 91\%  \\ \hline
Recall of \sceme           & 54\%  & 67\%  & 74\% & 89\% & 93\% \\ \specialrule{.1em}{.05em}{.05em}
\end{tabular}}}
\caption{Recall for detecting perturbed stop signs at different false positive rate.}
\label{tab:stop_sign}
\end{table}


\section{Conclusions}
Inspired by how humans can associate objects with where and how they appear within a scene, we propose to detect adversarial perturbations by recognizing context inconsistencies they cause in the input to a machine learning system.
We propose \sceme, which automatically learns four kinds of context, encompassing relationships within the scene and to the scene holistically.
Subsequently, we check for inconsistencies within these context types, and flag those inputs as adversarial.
Our experiments show that our method is extremely effective in detecting a variety of attacks on two large scale datasets and improves the detection performance by over 20\% compared to a state-of-the-art, context agnostic method.
\section*{Acknowledgments}
{
\color{black}{This research was partially sponsored by  ONR grant N00014-19-1-2264 through the Science of AI program, and
by the U.S. Army Combat Capabilities Development Command Army Research Laboratory under Cooperative Agreement Number W911NF-13-2-0045 (ARL Cyber Security CRA). The views and conclusions contained in this document are those of the authors and should not be interpreted as representing the official policies, either expressed or implied, of the Combat Capabilities Development Command Army Research Laboratory or the U.S. Government. The U.S. Government is authorized to reproduce and distribute reprints for Government purposes notwithstanding any copyright notation hereon.}
}
\bibliographystyle{splncs04}
\bibliography{newbib}
\newpage
\chapter*{Supplementary Material}
\appendix
\noindent
In this supplementary material, we provide: 
1) numbers used for the plots in the paper; 
2) the architecture of the auto-encoders;
3) how we extend the state-of-the-art adversarial perturbation detection method FeatureSqueeze to defend object detection system; 
4) how we apply non-deep Co-occurGraph to defend object detection system using cooccurrence relations inside the scene images; 
5) the detection performance of our proposed method against digital perturbations generated by various generation mechanisms;
6) comparing our proposed method with others that use context inconsistency to detect adversarial perturbations.
\section{Values in the Plots}
In the paper, some experimental results have been provided as plots for better visualization. We provide a table for each plot in this supplementary material.  Tab.~\ref{tab:cnt_proposal} and Tab.~\ref{tab:cnt_proposal_coco} correspond to the upper part and the lower part of Fig.8(a). Tab.~\ref{tab:iou} corresponds to Fig.8(b).
Tab.~\ref{tab:num_object} and Tab.~\ref{tab:num_object_coco} correspond to the upper part and lower part of Fig.8(c). Some entries are missing due to inadequate number of samples. For example, there are no entries for digital hiding attack for images with 6 objects in Tab.~\ref{tab:num_object} because there are only 14 hiding-attacked images and the AUC reported would not be accurate. We report AUC when we have at least 50 attacked samples.
\begin{table}[h!]
\centering
\resizebox{10.5cm}{!}{
\begin{tabular}{ccccccc}
\specialrule{.1em}{.05em}{.05em}
\multirow{2}{*}{\#Proposals} & \multicolumn{3}{c}{Digital Perturbations}            & \multicolumn{3}{c}{Physical Perturbations}           \\  
                                            & Miscategorization & Hiding & Appearing & Miscategorization & Hiding  & Appearing \\ \specialrule{.1em}{.05em}{.05em}
\textbf{FeatureSqueeze~\cite{xu2017feature}:} & & & & & &\\
\specialrule{.1em}{.05em}{.05em}                                           
1-3              & 0.751             & 0.628  & 0.587     & 0.762             & 0.679  & 0.647     \\ \hline
3-6              & 0.712             & 0.626  & 0.614     & 0.749             & 0.633  & 0.653     \\ \hline
6-9              & 0.748             & 0.612  & 0.609     & 0.784             & 0.654  & 0.688     \\ \hline
9-12             & 0.727             & 0.576  & 0.629     & 0.767             & 0.672  & 0.692     \\  \specialrule{.1em}{.05em}{.05em}
\textbf{Our method:} & & & & & &\\
\specialrule{.1em}{.05em}{.05em}
1-3              & 0.830             & 0.977  & 0.843     & 0.940             & 0.955  & 0.950     \\ \hline
3-6              & 0.902             & 0.995  & 0.859     & 0.983             & 0.982  & 0.977     \\ \hline
6-9              & 0.933             & 0.999  & 0.903     & 0.993             & 0.998  & 0.985     \\ \hline
9-12             & 0.950             & 0.983  & 0.929    & 0.996             & 1.000  & 0.991     \\  \specialrule{.1em}{.05em}{.05em}
\end{tabular}}
\caption{The detection performance against different attacks w.r.t. the number of proposals on the perturbed objects in PASCAL VOC dataset.}
\label{tab:cnt_proposal}
\vspace{-0.15in}
\end{table}

\begin{table}[h!]
\centering
\resizebox{10.5cm}{!}{
\begin{tabular}{ccccccc}
\specialrule{.1em}{.05em}{.05em}
\multirow{2}{*}{\#Proposals} & \multicolumn{3}{c}{Digital Attack}     & \multicolumn{3}{c}{Physical Attack}    \\ 
                             & Miscategorization & Hiding & Appearing & Miscategorization & Hiding & Appearing \\ \specialrule{.1em}{.05em}{.05em}
\textbf{FeatureSqueeze~\cite{xu2017feature}:} & & & & & &\\
\specialrule{.1em}{.05em}{.05em}
1-3                          & 0.704             & 0.692  & 0.594     & 0.670             & 0.678  & 0.502     \\ \hline
3-6                          & 0.656             & 0.679  & 0.569     & 0.719             & 0.692  & 0.528     \\ \hline
6-9                          & 0.584             & -      & 0.562     & 0.653             & 0.641  & 0.552     \\ \hline
9-12                         & 0.616             & -      & 0.521     & 0.682             & -      & 0.556     \\ \specialrule{.1em}{.05em}{.05em}
\textbf{Our method:} & & & & & &\\
\specialrule{.1em}{.05em}{.05em}
1-3                          & 0.896             & 0.961  & 0.804     & 0.918             & 0.938  & 0.952     \\ \hline
3-6                          & 0.947             & 0.992  & 0.876     & 0.985             & 0.982  & 0.973     \\ \hline
6-9                          & 0.978             & -      & 0.90      & 0.983             & 0.999  & 0.989     \\ \hline
9-12                         & 0.988             & -      & 0.932     & 0.995             & -      & 0.987     \\ \specialrule{.1em}{.05em}{.05em}
\end{tabular}}
\caption{The detection performance against different attacks w.r.t. the number of proposals on the perturbed objects in MS COCO dataset.}
\label{tab:cnt_proposal_coco}
\vspace{-0.15in}
\end{table}

\begin{table}[h!]
\centering
\resizebox{10.5cm}{!}{
\begin{tabular}{ccccc}
\specialrule{.1em}{.05em}{.05em}
\multirow{2}{*}{IoU} & \multicolumn{2}{c}{PASCAL VOC}                                                 & \multicolumn{2}{c}{MS COCO}                                                    \\ 
                     & \multicolumn{1}{l}{Digital Appearing} & \multicolumn{1}{l}{Physical Appearing} & \multicolumn{1}{l}{Digital Appearing} & \multicolumn{1}{l}{Physical Appearing} \\ \specialrule{.1em}{.05em}{.05em}
\textbf{FeatureSqueeze~\cite{xu2017feature}:} & & & &\\
\specialrule{.1em}{.05em}{.05em}
0.0                  & 0.605                                 & 0.653                                  & 0.614                                 & 0.550                                  \\ \hline
0.0-0.1              & 0.606                                 & 0.605                                  & 0.557                                 & 0.552                                  \\ \hline
0.1-0.2              & 0.592                                 & 0.642                                  & 0.549                                 & 0.518                                  \\ \hline
0.2-0.3              & 0.602                                 & 0.752                                  & 0.521                                 & 0.478                                  \\ \hline
0.3-0.4              & 0.590                                 & 0.640                                  & 0.504                                 & 0.586                                  \\ \hline
0.4-0.5              & 0.594                                 & 0.644                                  & 0.510                                 & 0.474                                  \\ \specialrule{.1em}{.05em}{.05em}
\textbf{Our method:} & & & & \\
\specialrule{.1em}{.05em}{.05em}
0.0                  & 0.748                                 & 0.939                                  & 0.769                                 & 0.977                                  \\ \hline
0.0-0.1              & 0.872                                 & 0.945                                  & 0.827                                 & 0.970                                  \\ \hline
0.1-0.2              & 0.879                                 & 0.966                                  & 0.849                                 & 0.978                                  \\ \hline
0.2-0.3              & 0.906                                 & 0.980                                  & 0.850                                 & 0.984                                  \\ \hline
0.3-0.4              & 0.905                                 & 0.986                                  & 0.855                                 & 0.996                                  \\ \hline
0.4-0.5              & 0.924                                 & 0.994                                  & 0.910                                 & 0.990                                  \\ \specialrule{.1em}{.05em}{.05em}
\end{tabular}}
\caption{The detection performance against appearing attacks w.r.t. the overlap (IoU) between the perturbed region and some ground truth object in PASCAL VOC and MS COCO}
\label{tab:iou}
\vspace{-0.15in}
\end{table}

\begin{table}[h!]
\centering
\resizebox{10.5cm}{!}{
\begin{tabular}{ccccccc}
\specialrule{.1em}{.05em}{.05em}
\multirow{2}{*}{\#Objects} & \multicolumn{3}{c}{Digital Perturbation}            & \multicolumn{3}{c}{Physical Perturbations}           \\ 
                            & Miscategorization & Hiding & Appearing & Miscategorization & Hiding & Appearing \\ \specialrule{.1em}{.05em}{.05em}
\textbf{FeatureSqueeze~\cite{xu2017feature}:} & & & & & &\\
\specialrule{.1em}{.05em}{.05em}
1                           & 0.724             & 0.627  & 0.600     & 0.726             & 0.617  & 0.657     \\ \hline
2                           & 0.715             & 0.624  & 0.574     & 0.806             & 0.679  & 0.635     \\ \hline
3                           & 0.733             & 0.610  & 0.661     & 0.834             & 0.716  & 0.631     \\ \hline
4                           & 0.760             & 0.615      & 0.584     & 0.806             & 0.683  & 0.578     \\ \hline
5                           & 0.740             & 0.612      & 0.611     & 0.879             & 0.789  & 0.640     \\ \hline
6                           & 0.778             & -      & 0.666     & 0.825             & 0.735  & 0.675     \\ \specialrule{.1em}{.05em}{.05em}
\textbf{Our method:} & & & & & &\\
\specialrule{.1em}{.05em}{.05em}1                           & 0.927             & 0.994  & 0.829     & 0.986             & 0.987  & 0.966     \\ \hline
2                           & 0.901             & 0.986  & 0.838     & 0.972             & 0.940  & 0.979     \\ \hline
3                           & 0.888             & 0.960  & 0.810      & 0.913             & 0.898  & 0.977     \\ \hline
4                           & 0.889             & 0.969  & 0.813     & 0.984             & 0.976  & 0.987     \\ \hline
5                           & 0.890             & 0.958  & 0.902     & 0.980             & 1.000  & 0.986     \\ \hline
6                           & 0.912             & -      & 0.968     & 0.987             & 0.998  & 0.995     \\ \specialrule{.1em}{.05em}{.05em}
\end{tabular}}
\caption{The detection performance against different attacks w.r.t. the number of objects in the scene images in PASCAL VOC dataset.}
\label{tab:num_object}
\vspace{-0.15in}
\end{table}

\begin{table}[h!]
\centering
\resizebox{10.5cm}{!}{
\begin{tabular}{ccccccc}
\specialrule{.1em}{.05em}{.05em}
\multirow{2}{*}{\#Object} & \multicolumn{3}{c}{Digital Attack}     & \multicolumn{3}{c}{Physical Attack}    \\ 
                          & Miscategorization & Hiding & Appearing & Miscategorization & Hiding & Appearing \\ \specialrule{.1em}{.05em}{.05em}
\textbf{FeatureSqueeze~\cite{xu2017feature}:} & & & & & &\\
\specialrule{.1em}{.05em}{.05em}
1                         & 0.683             & 0.681  & 0.590      & 0.674             & 0.701  & 0.565     \\ \hline
2                         & 0.677             & 0.676  & 0.573     & 0.692             & 0.688  & 0.550     \\ \hline
3                         & 0.693             & 0.683  & 0.562     & 0.714             & 0.636  & 0.539     \\ \hline
4                         & 0.676             & 0.691  & 0.584     & 0.707             & 0.749  & 0.532     \\ \hline
5                         & 0.662             & 0.676  & 0.528     & 0.654             & 0.596  & -         \\ \hline
6                         & 0.699             & 0.683  & 0.611     & 0.751             & 0.621  & -         \\ \specialrule{.1em}{.05em}{.05em}
\textbf{Our method:} & & & & & &\\
\specialrule{.1em}{.05em}{.05em}1                         & 0.976             & 0.991  & 0.853     & 0.993             & 0.957  & 0.984     \\ \hline
2                         & 0.964             & 0.987  & 0.824     & 0.984             & 0.967  & 0.975     \\ \hline
3                         & 0.922             & 0.972  & 0.884     & 0.982             & 0.967  & 0.967     \\ \hline
4                         & 0.891             & 0.938  & 0.882     & 0.986             & 0.984  & 0.936     \\ \hline
5                         & 0.952             & 0.963  & 0.903     & 0.995             & 0.992  & 0.995     \\ \hline
6                         & 0.965             & 0.983  & 0.909     & 0.991             & 0.994  & 0.997     \\ \specialrule{.1em}{.05em}{.05em}
\end{tabular}}
\caption{The detection performance against different attacks w.r.t. the number of objects in the scene images in COCO dataset.}
\label{tab:num_object_coco}
\vspace{-0.15in}
\end{table}

\section{Architecture of the Auto-encoders}
For each category, we use a separate auto-encoder to learn the distribution of its context profile. The architecture of the auto-encoders is identical and is shown in Fig.~\ref{fig:auto-encoder}.
The input to the auto-encoder is the context profile $x=[r,\gamma_{u1},\gamma_{u2},\gamma_{r1},\gamma_{r2}]$. We denote the height and width of the input as $H$ and $W$. $W=5$ since there are 5 feature vectors in $x$ and $H$ equals to the dimension of the RoI pooling feature.
A fully connected layer is first used to compress the node features ($r$) and edge features ($[\gamma_{u1},\gamma_{u2},\gamma_{r1},\gamma_{r2}]$) separately.
This is followed by two convolution layers, wherein the node and edge features are combined to learn the joint compression.
Two fully connected layers are then used to further compress the joint features.
These layers form a bottleneck that drives the encoder to learn the true relationships between the features and get rid of redundant information.

\begin{figure}[t]
    \begin{center}
        \parbox{4.6in} {
            \centerline{ \includegraphics[width=9cm]{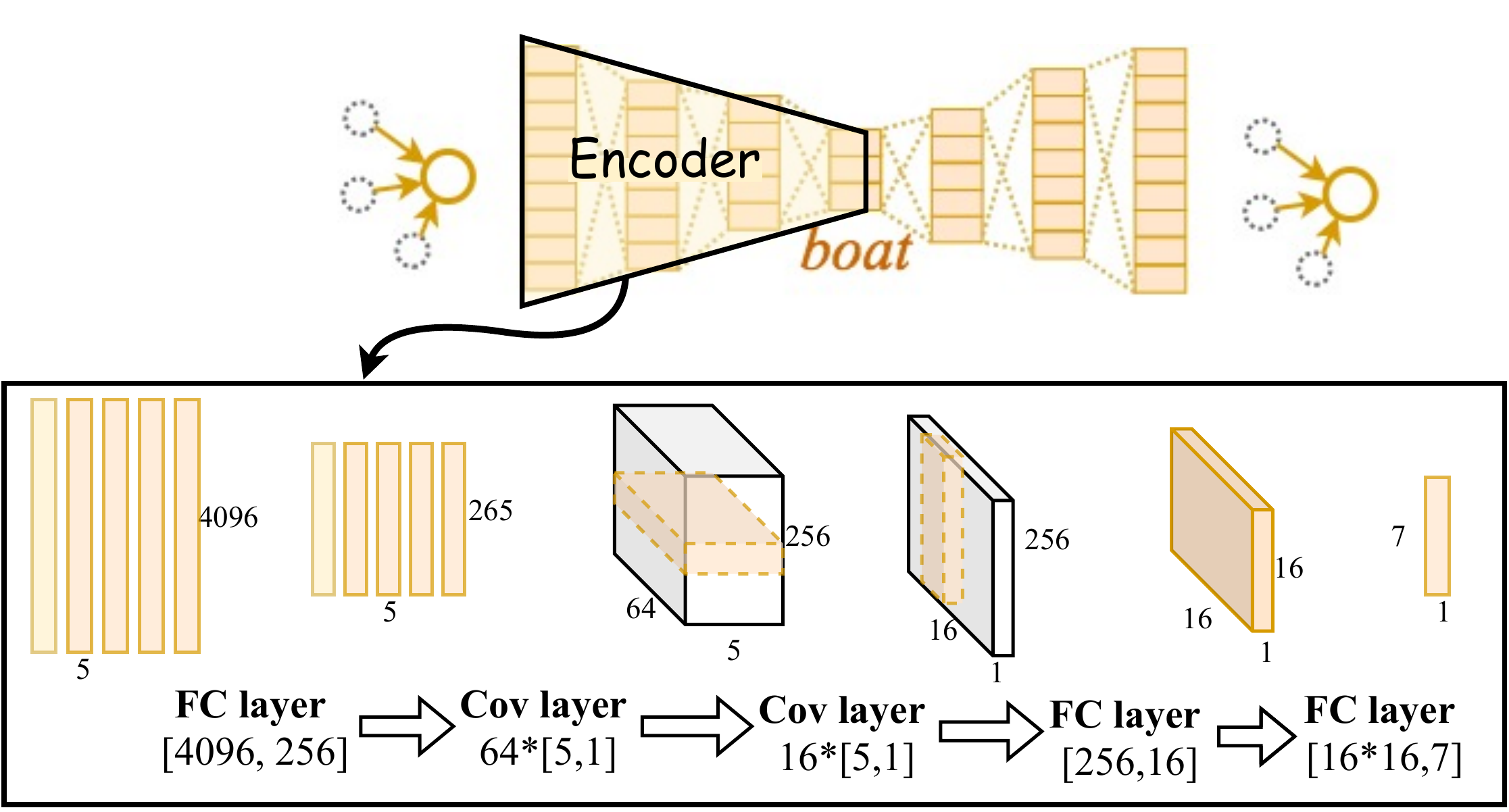}}
            \caption{{ Auto-encoder structure.} One auto-encoder is learned for each category. The structure of the auto-encoders is identical.}
            \label{fig:auto-encoder}}
\end{center}
\end{figure}

\section{Extending FeatureSqueeze to Region-level Perturbation Detection}
\subsection{FeatureSqueeze}
FeatureSqueeze~\cite{xu2017feature} proposes to squeeze the search space available to an adversary, driven by the observation that the feature input spaces are often unnecessarily large, which provides extensive opportunities for an adversary to construct adversarial examples. There are two feature squeezing methods used in their implementation: a) reducing the color bit depth of each pixel; b) spatial smoothing. By comparing a DNN model's prediction on the original input with that on squeezed ones, feature squeezing detects adversarial examples with high accuracy and few false positives. The framework of FeatureSqueeze~\cite{xu2017feature}  is shown in Fig.~\ref{fig:FeatureSqueeze}.

\begin{figure}[t!]
        \begin{center}
                \parbox{4.6in} {
                        \centerline{ \includegraphics[width=10.5cm]{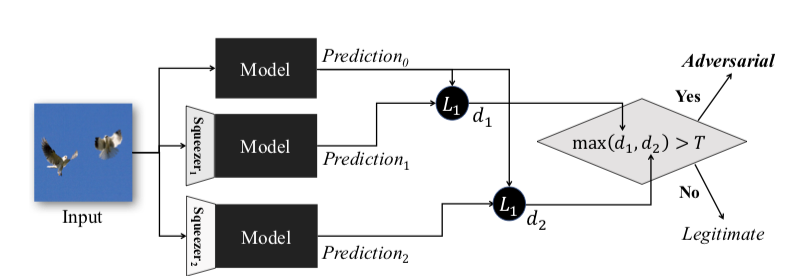}}
                        \caption{ This figure is from  paper~\cite{xu2017feature}. ``The model is evaluated on both the original input and the input after being pre-processed by feature squeezers. If the difference between the model’s prediction on a squeezed input and its prediction on the original input exceeds a threshold level, the input is identified to be adversarial.''}
                        \label{fig:FeatureSqueeze}}
\vspace{-0.35in}
\end{center}
\end{figure}

\begin{figure}[t!]
        \begin{center}
                \parbox{4.6in} {
                        \centerline{ \includegraphics[width=10.5cm]{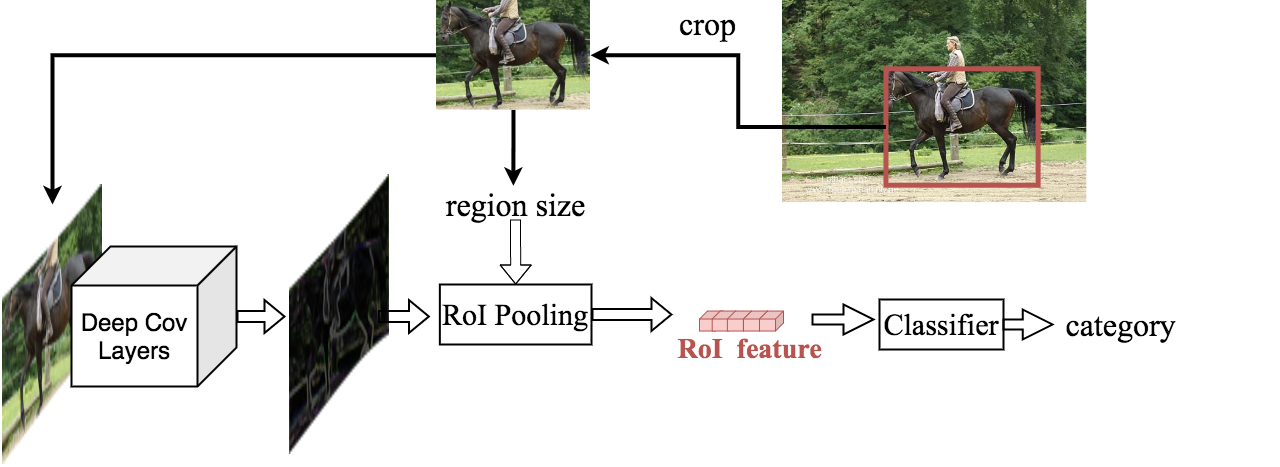}}
                        \caption{Extending the DNN of FeatureSqueeze to region-level classification}
                        \label{fig:ExtendFeatureSqueeze}}
\vspace{-0.35in}
\end{center}
\end{figure}

\subsection{Extending to Region-level Detection}
To detect perturbed regions inside scene images, the DNN model of FeatureSqueeze is required to operate on region-level. We crop the ground-truth regions, denoted as $r$, as the input to the DNN model. The output of the DNN model is the predicted category. To deal with region inputs with various size, we use RoI pooling~\cite{girshick2015fast} (box size equals to input region size) as the last feature extraction layer as shown in Fig.~\ref{fig:ExtendFeatureSqueeze}. Softmax function~\cite{bishop2006pattern} is used as the last layer and cross entropy loss~\cite{goodfellow2016deep} is used as the objective loss function.
\subsection{Implementation Details}
We initialize the feature extractor with the weights pretrained on ImageNet. Momentum optimizer with momentum 0.9 is used to train the classifier. The learning rate is 1e-4 and decays every 80k iterations at decay rate 0.1. Training ends after 240k iterations. The final classification accuracy for the 20 categories in PASCAL VOC dataset is 95.6\%.  The final classification accuracy for the 80 categories in MS COCO dataset is 87.1\%. The accuracy is not high because MS COCO is biased among categories, for example, more than 100k person instances v.s. less than 1k hair dryer instances. Even after we balance the number of samples among different categories, the performance is not good because some categories have too few examples, like the hair dryer category.
The hyperparameters used for feature squeezing are exactly the same as the authors' GitHub implementation~\cite{uvasrg}. 

\section{Co-occurGraph for Misclassification Attack Detection} 

We consider a non-deep model as baseline where co-occurrence statistics are used to detect misclassification due to adversarial perturbation. This approach uses the inconsistency between prior relational information obtained from the training data and inferred relational information conditioned on misclassified detection to detect the presence of adversarial perturbation. As the co-occurrence statistics of background class cannot be modeled, this approach is not applicable for detecting hiding and appearing attacks.

\noindent
\textbf{Prior Relational Information.} Same as \cite{bappy2016online}, we use the co-occurrence frequency of different categories of objects in the training data to obtain the prior relational information. Co-occurrence statistics gives an estimate of how likely two object classes will appear together in an image.  

\noindent
\textbf{Graphical Representation.} To encode the relational information of different classes of objects present in an image, we represent each image as an undirected graph $G=(V, E)$. Here, a node in $V$ represents a single proposed region by the region proposal network. The edges $E=\{(i, j)|$ if region $v_i$ and $v_j$ are linked$\}$ represent the relationships between the regions. We formulate a tree structure graph where the region of interest is connected with all other proposed regions. The estimate of class probabilities of each proposed region generated by the object detection model is used as the node potential and the co-occurrence statistics is used as the edge potential.  

\noindent
\textbf{Detection of Misclassification Attack.} For each image instance in test-set, we estimate its class conditional relatedness with other classes by making conditional inference on the representative graph. Conditional inference gives the pairwise conditional distribution of classes for each edge, which we use to obtain the posterior relational information of that image conditioned on the misclassified label. Based on the inconsistency among the prior relational information and  posterior relational information, we detect if there is any misclassification attack.

\noindent
\textbf{Implementation Details.} We use the Faster R-CNN \cite{ren2015faster} as the object detection and region proposal generation module. For each image, we consider top 20 proposed regions based on the class confidence score. To formulate the graph and make conditional inference, we use the publicly available UGM Toolbox \cite{schmidt2012ugm}.
\section{Detection performance w.r.t. various perturbation generation mechanisms}
In the paper, we show our proposed method is effective in detecting  six different perturbation attacks, i.e., digital miscategorization attack, digital hiding attack, digital appearing attack, physical miscategorization attack, physical hiding attack and physical appearing attack. These attacks are different in terms of their attack goals and perturbation forms.
Other defense papers also evaluate their defense methods w.r.t different perturbation generation mechanisms. Our defense strategy is dependent on the contextual information, and therefore should not rely heavily on the mechanism to
generate the perturbation. We validate our hypothesis by testing our method against different perturbation generation mechanisms. The results in Tab.~\ref{tab:other_attacks} show that our method is consistently effective against all the perturbation generation mechanisms.

As stated in the paper, COCO has few examples for certain categories. To make sure we have enough number of context profiles to learn the distribution, out of all the 80 categories, we choose 10 categories with the largest number of context profiles extracted. These 10 categories are ``car'', ``diningtable'', ``chair'', ``bowl'', ``giraffe'', ``person'',  ``zebra'', ``elephant'', ``cow'', ``cat''. We also choose ``stop sign'' category because attacks on stop signs have gained long-lasting attentions. In addition to ``background'', we have in total 12 categories and learn 12 autoencoders separately. We use these 12 autoencoders and evaluate misclassifications to these categories in our experiments.
\begin{table}[h!]
\centering
\resizebox{8.5cm}{!}{
\setlength{\tabcolsep}{3mm}{
\begin{tabular}{lcc}
\specialrule{.1em}{.05em}{.05em}
{Perturbation Generation Mechanism} &{PASCAL VOC}                                                 & {MS COCO} \\ \specialrule{.1em}{.05em}{.05em}
\textbf{FeatureSqueeze~\cite{xu2017feature}:} & & \\
\specialrule{.1em}{.05em}{.05em}

FGSM~\cite{goodfellow2014explaining} &0.788 &0.678\\ \hline
BIM~\cite{kurakin2016adversarial} &0.724 &0.681\\ \hline
\textbf{Our method:} & & \\
\specialrule{.1em}{.05em}{.05em}
FGSM &0.947 &0.915\\ \hline
BIM &0.938 &0.959\\ \hline
\end{tabular}}}
\caption{The detection performance against digital miscategorization attacks w.r.t. different perturbation generation mechanisms on PASCAL VOC and MS COCO}
\label{tab:other_attacks}
\vspace{-0.15in}
\end{table}

\section{Comparison with other context inconsistency based adversarial defense methods}
The general notion of using context
has been used to detect anomalous activities\cite{zhu2012context,xu2014video,cao2017voila,hasan2016learning}. When it comes to adversarial perturbation detection, spatial context has been used to detect adversarial perturbations against semantic segmentation~\cite{xiao2018characterizing}. Temporal context  has been used to detect adversarial perturbation against video classification~\cite{jia2019identifying}. Context inconsistency has never been used to
detect adversarial examples against objection detection systems. 
Essentially, our approach utilizes different kinds of context, including the spatial one from
these prior works and object-level inter-relationships for the first time, as discussed in Tab.~\ref{tab:comparison}.
\begin{table}[h!]
\centering
\resizebox{12cm}{!}{
\begin{tabular}{c|c|c|c|c|c|c}
\specialrule{.1em}{.05em}{.05em}
Detection                     & Temporal & Spatial  & Object-object  & Object-background  & Object-scene & Task                  \\ \specialrule{.1em}{.05em}{.05em}
Video\cite{jia2019identifying}     & \cmark           &                 &                       &                           &                      & video classification  \\ \hline
Seg\cite{xiao2018characterizing} &                  & \cmark          &                       &                           &                      & semantic segmentation \\ \hline
Our method                    &                  & \cmark          & \cmark                & \cmark                    & \cmark               & object detection      \\ \specialrule{.1em}{.05em}{.05em}
\end{tabular}}
\caption{Comparison with other context inconsistency based adversarial detection methods}
\label{tab:comparison}
\end{table}
\end{document}